\documentclass[a4paper,11pt]{article}

\newcommand{\STAB}[1]{\begin{tabular}{@{}c@{}}#1\end{tabular}}
\usepackage{color}
\usepackage{multicol}
\usepackage{multirow}
\usepackage{comment}
\usepackage{fourier}
\usepackage{color}
 \usepackage{graphicx}
\usepackage{url}
\usepackage[affil-it]{authblk}
\usepackage{amsmath}
\usepackage{wrapfig}
\usepackage{xspace}
\usepackage[utf8]{inputenc}

\usepackage[T1]{fontenc}
\usepackage{times}

\begin{document}

\title{Multi-Person tracking by multi-scale detection in Basketball scenarios}

\author{Adrià Arbués-Sangüesa$^{1,2}$, Gloria Haro$^1$, Coloma Ballester$^1$}
\affil{$^1$ Universitat Pompeu Fabra (Barcelona, Spain) and $^2$ FC Barcelona}
\date{}
\maketitle
\thispagestyle{empty}

\begin{abstract}
Tracking data is a powerful tool for basketball teams in order to extract advanced semantic information and statistics that might lead to a performance boost. However, multi-person tracking is a challenging task to solve in single-camera video sequences, given the frequent occlusions and cluttering that occur in a restricted scenario. In this paper, a novel multi-scale detection method is presented, which is later used to extract geometric and content features, resulting in a multi-person video tracking system. Having built a dataset from scratch together with its ground truth (more than 10k bounding boxes), standard metrics are evaluated, obtaining notable results both in terms of detection (F1-score) and tracking (MOTA). The presented system could be used as a source of data gathering in order to extract useful statistics and semantic analyses \textit{a posteriori}.\\
\end{abstract}

\section{Introduction}
\begin{figure}[h]
\centerline{\includegraphics[width=\linewidth]{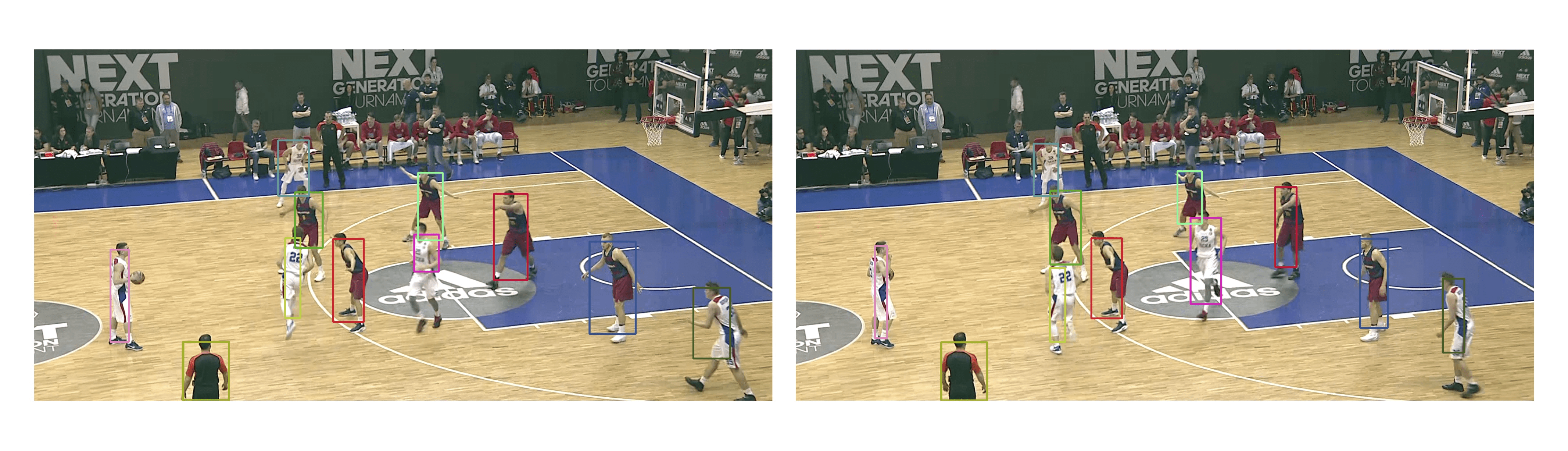}}
\caption{Obtained results in adjacent frames, where all players (and referee) in court are properly detected (bounding box) and tracked (color identifier).}
\label{fig:boxes}
\end{figure}

Right in the \textit{Big Data Era}, sports teams are gathering a lot of advanced statistics about players to make the appropriate decisions that may lead to a boost of performance; for instance, hiring a new player or designing optimal tactics for a specific game. In particular, in the basketball field, a strong demand for tracking data has emerged, and video-based companies such as Second Spectrum 
or STATS 
extract this information with a set of overhead cameras placed at the ceiling of the stadiums. However, at the moment, there is not an established implemented way to infer tracking statistics from simple single-camera video sequences, due to the many challenges that emerge, such as multiple occlusions, similarity in appearance or fast and erratic motions \cite{thomas2017computer}. Besides, the accurate detection and tracking of persons in a video is a pivotal issue for the automatic analysis and understanding of the actions happening in the scene captured by the camera.In this article, a novel multi-tracker method thought for single-camera basketball sequences is presented, where all targets are properly tracked with 0.67 MOTA confidence; sample results are shown in Figure \ref{fig:boxes}.

\section{State of the Art}

There has been a considerable effort towards multi-object tracking in video. Frequently, the tracking problem is approached by a previous or simultaneous detection step.
Person detection and tracking is used in \cite{ramanathan2016detecting} to detect events in multi-person videos. The authors propose to use a CNN-based multibox detector \cite{szegedy2013deep} and a KLT tracker \cite{veenman2001resolving}. A human pose estimation method that extends the Mask R-CNN \cite{he2017mask} to the video case is presented in \cite{girdhar2018detect}, which includes a tracking cost based on three terms: Intersection over Union (IoU) of the bounding boxes, a pose similarity metric, and a similarity metric that uses CNN features.
An optimization method for the joint segmentation and tracking of scenes with multiple targets is proposed in \cite{milan2015joint}. The assignment of part of object (or superpixel) detections to trajectory hypothesis is formulated as a multi-label conditional random field. 
\cite{henschel2018fusion} 
propose a method that uses two different detectors, namely, a full-body detector and a head detector. All detections are incorporated in a global optimization formulation which translates in a binary quadratic problem that is solved by relaxation through the Frank-Wolfe algorithm. \cite{doering2018joint} propose a method 
based on a Siamese network that evaluates the person pose in two frames at a time and a temporal CNN that predicts the so-called Temporal Flow Fields, 
a graph optimization problem to obtain the tracking. 
Pose tracking refers in the literature to the task of estimating anatomical human keypoints and assigning unique labels for each  keypoint across the frames of a video \cite{iqbal2017posetrack,insafutdinov2017arttrack}. 
Our method for detection and tracking estimates the human pose in each frame and uses the similarity of the detected pose keypoints to reinforce the tracking of a given person. Our focus is on team sports, and in particular, basketball games. For a thorough revision of computer vision techniques applied to the analysis of sports (including players detection and tracking) we refer the reader to the relatively recent survey \cite{thomas2017computer}. In a recent work and for the basketball case,  \cite{senocak2018part} propose a deep learning based method for player identification that incorporates features on body parts which are in turn computed with Convolutional Pose Machines \cite{poseModel2}.

\section{Proposed method}
\label{sec:method}
\begin{figure}[h]
\centerline{\includegraphics[width=0.8\linewidth]{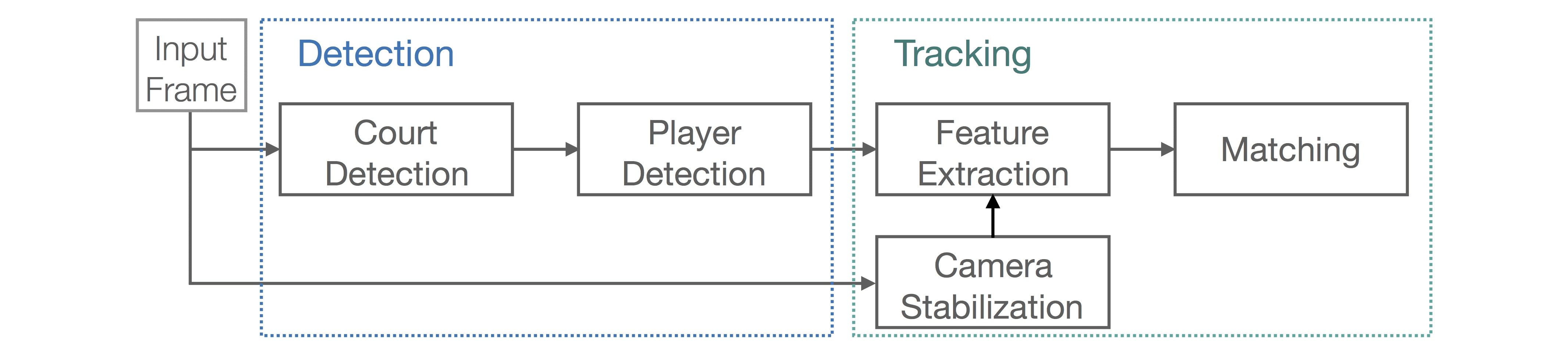}}
\caption{Overall pipeline of the presented method.}
\label{fig:pipeline}
\end{figure}

Our proposal to track players in a basketball game is based on a tracking-by-detection approach as seen in Figure \ref{fig:pipeline}: First, the different individuals on the court are detected in every frame and then, matches are established along time. Given a video of a basketball game, the prior that  players are inside the court along the sequence is used to restrict the area where players should be searched, thus avoiding the detection of spectators or bench players. Once this region is detected, players on court are found and, having stabilized the video sequence, geometric and content features are extracted for every single detected instance. Having matched the different detections from adjacent frames, the tracking of players along the video is obtained.

\subsection{Court Detection}
The court is a rectangular area whose projection to the camera results in a trapezoid. Thus, the problem is reduced to the identification of visible court boundaries in the image: the sidelines and baselines (from 1 to 4 depending on the camera the point of view). Frequently, some of these court boundaries are only partially visible due to occlusions produced by, \textit{e.g}., players, referees, the public and the synthetic scoreboard.  

\begin{figure}[t]
\centerline{\includegraphics[width=0.6\linewidth]{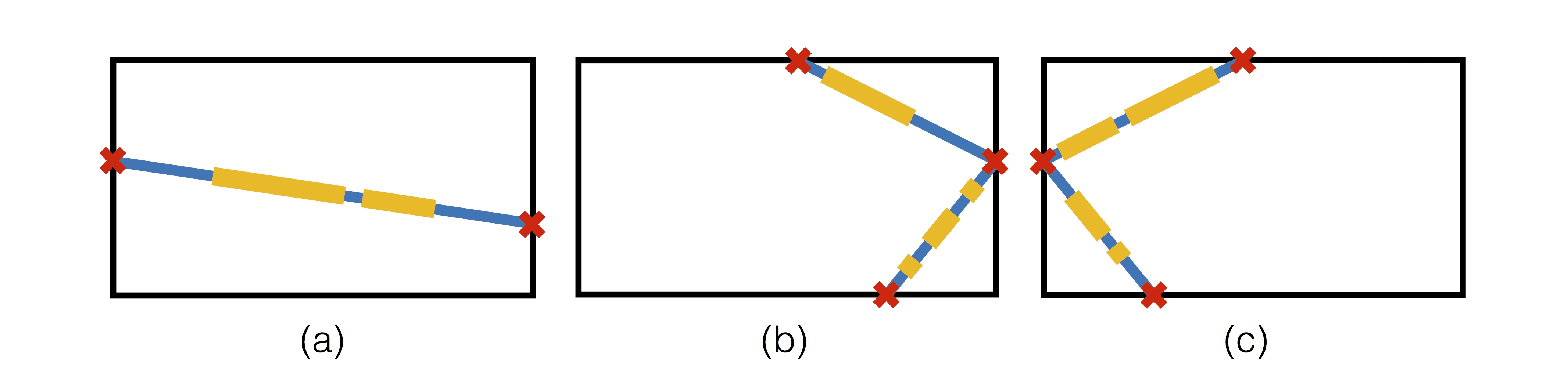}}
\caption{Line contributions. (a) potential sidelines to be detected. (b)-(c) right-left baselines, respectively.}
\label{fig:courtDet1}
\end{figure}

\begin{figure}[t]
\centerline{\includegraphics[width=\linewidth]{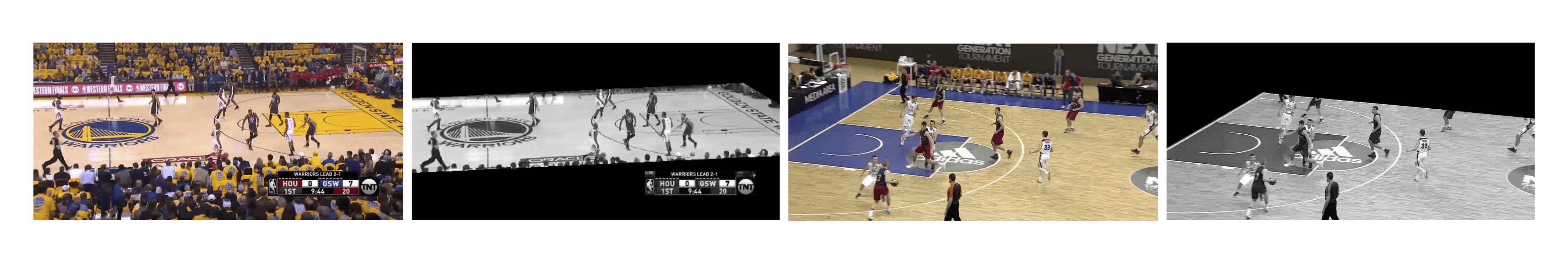}}
\caption{Court detection results in different scenarios: (left) NBA, and (right) European games}
\label{fig:courtDet2}
\end{figure}

The method starts by detecting all the line segments in the image using a fast and robust parameter-less method \cite{von2010lsd}. Right after, dominant lines, \textit{i.e.} lines the with longest visible parts, are estimated using a voting 
procedure. Those lines will correspond, in general, to the sidelines/baselines or, in cases of strong occlusions, to court lines parallel to the sidelines/baselines. 
The strength of the vote of each line is proportional to the sum of 
detected segments' length on the line, as seen in Fig.~\ref{fig:courtDet1}, where the detected segments are shown in yellow. 
Given that in broadcasting sequences only one baseline (or none) can be seen at a time and that even in cases that both sidelines are in the field of view of the camera one of them may appear occluded by the public (\textit{e.g.} Fig.~\ref{fig:courtDet2}), 
the purpose is to find a \emph{horizontal} dominant line (either a sideline or its orientation) and a \emph{vertical} dominant one (a baseline or its orientation). Horizontal lines are considered to be the ones which intersect the image at the left and right boundaries (Fig.~\ref{fig:courtDet1}(a)), while vertical ones intersect in one of the following pairs of image sides: top-left, bottom-left, top-right or bottom-right (examples 
in Fig.~\ref{fig:courtDet1}(b)-(c)). 
In order to find the location of court boundaries, the court is pre-segmented and the set of lines (with the orientation of the horizontal or vertical dominant lines) that better delimits the pre-segmented court is selected. 
Two different solutions are proposed to pre-segment the court in two different professional basketball scenarios: (a) European, and (b) NBA games. \\ 
In European games (Fig.~\ref{fig:courtDet2}-right), court surroundings usually share the same color, and fans sit far from team benches. For this reason, a basic color filter (in the HSV colorspace) can be created; for each possible line candidate, the contribution of pixels that satisfy filter conditions is checked at both right-left (vertical) or above-below (horizontal) sides of the tested candidate. The horizontal and vertical candidates with the highest response in terms of difference will be then considered as court limits. \\
For NBA games (Fig.~\ref{fig:courtDet2}-left), the scenario is much more challenging, because there is almost no space between sidelines and fans. In order to find the horizontal boundaries, instead of checking for color components, Conditional Random Fields \cite{Zheng2015} is applied at a coarse resolution to find the total area of people pixels; this estimation is later complemented with Histogram of Oriented Gradients. Once having this rough estimation, an iterative algorithm is applied to delimit court boundaries: at the very beginning, two line candidates with the dominant orientation are placed at the top and bottom of the image; then, for each iteration, these lines are moved towards the middle until convergence. In each iteration, the product of the following percentages is computed: (a) people-pixels above the top line, (b) people-pixels below the bottom line, and (c) non-people-pixels below the top line and above the bottom one. If there is a drop in the first or second percentage, the position of the corresponding line is fixed; convergence is reached when both lines stop moving. Potentially, in the horizontal court limits, the product of these three terms will correspond to a maximum, meaning that there is a large contribution of people pixels above and below the top and bottom line respectively (corresponding to fans), and a small contribution in between (corresponding to the court with a maximum of 10 players plus 3 officials). Having masked the original image given the detected sidelines, the best vertical candidate is found in the same way but scanning only from left to right / right to left.

%
 




\subsection{Player Detection}\label{subsec:playerdect}
Our detection method is based on a robust strategy that relies on pose models techniques \cite{poseModel1,poseModel2,cao2017realtime}. 
Stemming from a TensorFlow implementation of OpenPose\footnote{\url{https://github.com/ildoonet/tf-pose-estimation}} thought for low-powered embedded devices, performance on high resolution videos is improved by a multi-scale strategy that refines the detection at the original training scale. More precisely, the method of \cite{cao2017realtime} is a bottom-up approach that identifies up to 17 anatomical keypoints (corresponding to parts of the face, body, etc) and joins them in limbs connecting two keypoints and, finally, in the visible person skeleton. It consists on a CNN that incorporates features given by a VGG-19 network and obtains part confidence maps and a nonparametric representation which is referred to as Part Affinity Fields encoding location and orientation of the limbs. The authors use the part confidence maps together with the part affinity fields to extract the person skeleton.

\begin{figure}[t]
\centerline{\includegraphics[width=0.7\linewidth]{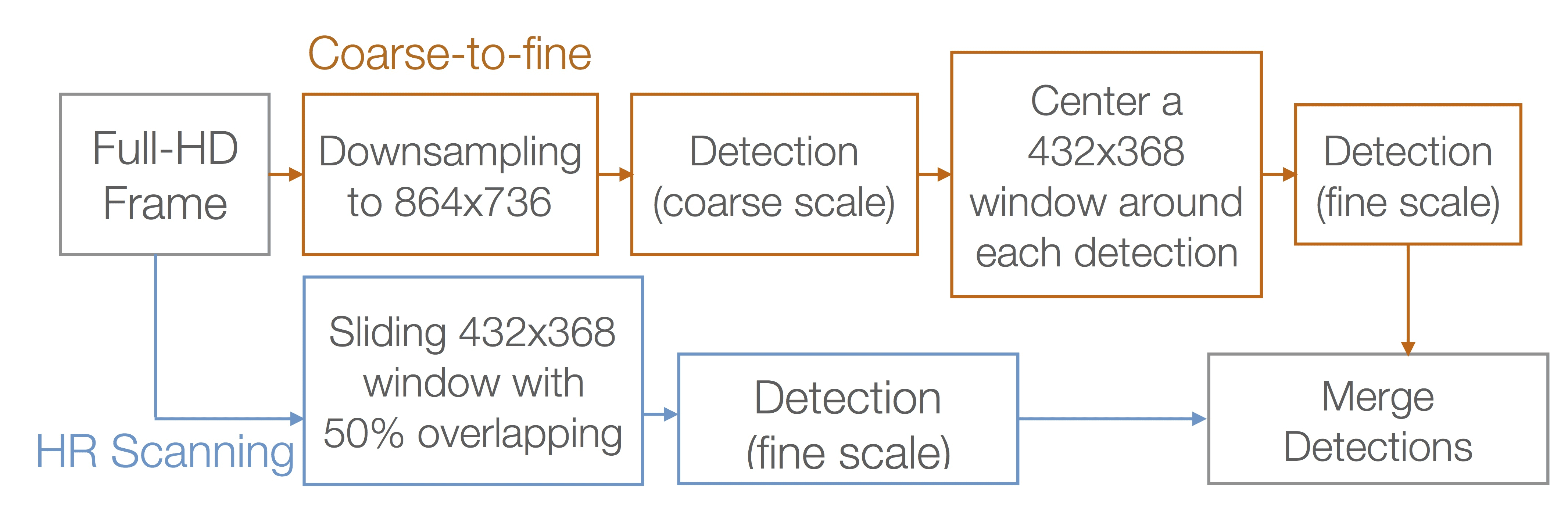}}
\caption{Proposed multi-scale detection strategy.}
\label{fig:combHD}
\end{figure}

The model being used in the above-mentioned OpenPose implementation was trained with a MobileNet network, with a default image resolution of 432 $\times$ 368 pixels, and it has been experimentally checked that best results are obtained at that resolution. With the purpose of maximizing the accuracy of human pose detection in our full-HD frames of 1920 $\times$ 1080 pixels, the following multi-scale strategy (Figure \ref{fig:combHD}) is proposed: \\
\textbf{1A}: The frame is downscaled to twice the resolution of the pretrained model images, and pose estimation is applied. Most of the players are detected but their pose is not accurate enough due to the loss of resolution. \\
\textbf{1B}: Detections at the coarse scale are refined by centering a 432$\times$368 window around the center of each former detection in the original HR image. The human pose is better estimated and new detections may emerge, specially those corresponding to players partially occluded by other players, with only small visible areas. \\
\textbf{2}: To include those players who have not been detected at the first low-resolution stage (typically corresponding to players with motion blur who even fade away more after downscaling), the detection is carried out with a sliding window of size 432$\times$368 over the full-HD image, with 50\% overlapping, both horizontally and vertically. Detections not found at the current stage are added to the former ones. 

Notice that the presented person detection method does not include priors such as a maximum number of players in the basketball court or information about the team uniforms and, thus, the set of detected persons might include some referees (an example is displayed in Fig.~\ref{fig:boxes}). 
The detection output is a bounding box placed in the player skeleton position (Fig.~\ref{fig:opBoxes} shows some examples), and a heatmap of 18 layers (one for each part, plus their combination) for each detection, indicating the confidence of each part being at each particular pixel. 

\subsection{Multi-Person Tracking}
Before tracking the individual players we propose to remove the camera motion by using a camera stabilization method  \cite{sanchez2017ipol}, which outputs a set of homographies $\{ H_t\}_t$, being $H_t$ the matrix that stabilizes the frame at time $t$. For this reason, it can be assumed that, in general, the player motion within frames is small. 

\begin{figure}[t]
\centerline{\includegraphics[width=0.7\linewidth]{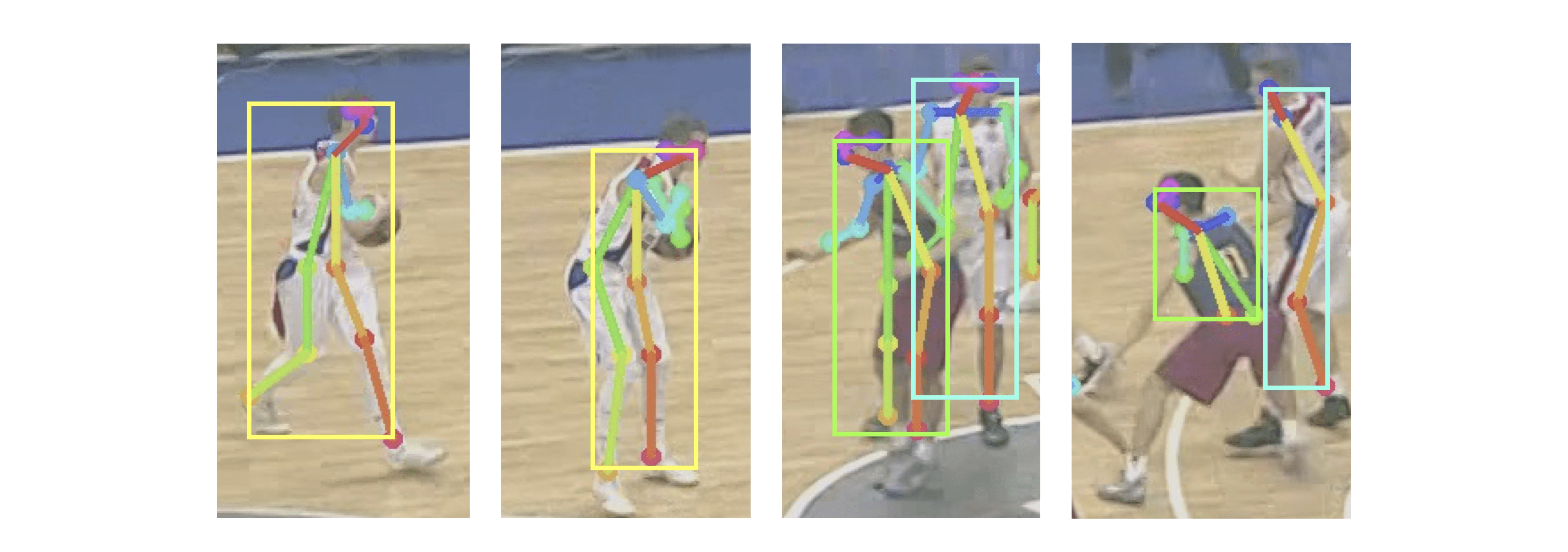}}
\caption{Detected parts with the corresponding bounding box.}
\label{fig:opBoxes}
\end{figure}

In order to track the bounding boxes corresponding to the players (or the detected part of them, as seen in Figures \ref{fig:boxes} and \ref{fig:opBoxes}), a similarity cost is defined, which is made of three terms: two geometric terms and a content-based one. Let us denote by $I_{t_1}$ and $I_{t_2}$ two frames of the input video. 
Stabilized versions are considered by $H_{t_1}$ and $H_{t_2}$, respectively. Given two bounding boxes $B_{t_1}$ and $B_{t_2}$, detected in $I_{t_1}$ and $I_{t_2}$, respectively, the proposed similarity cost is defined as:
\begin{equation}
C(B_{t_1},B_{t_2})= \alpha {C}_{d}(B_{t_1},B_{t_2}) + \beta {C}_{i}(B_{t_1},B_{t_2}) + \gamma {C}_{c}(B_{t_1},B_{t_2}),
\label{eq:Cost1}
\end{equation}
where $\alpha,\beta\in [0,1]$, $\gamma=1-(\alpha+\beta)$, and $C_{d},C_{i},C_{c}$ in Equation \eqref{eq:Cost1} are defined as follows:

\noindent
{\bf (a)} $C_{d}(B_{t_1},B_{t_2})$ is the normalized distance between the transformation by $H_{t_1}$ and $H_{t_2}$, respectively, of the centroids of the bounding boxes, $\mathbf{x}_{B_{t_1}}$ and $\mathbf{x}_{B_{t_2}}$. That is, $C_d(B_{t_1},B_{t_2})=\frac{1}{\sqrt{w^2+h^2}}\|H_{t_1}(\mathbf{x}_{B_{t_1}})-H_{t_2}(\mathbf{x}_{B_2})\|$, where $w$ and $h$ are  the width and the height of the frame domain. 

\noindent
{\bf (b)} $C_{i}(B_{t_1},B_{t_2})=\text{IoU}(H_{t_1}\left(B_{t_1}),H_{t_2}(B_{t_2})\right)$ is the Intersection over Union (IoU) value of the transformation by $H_{t_1}$ and $H_{t_2}$ of the two bounding boxes, respectively.

\noindent    
{\bf (c)} The content of $B_{t_1}$ and $B_{t_2}$ is compared by considering only the pairs of anatomical keypoints (joints between limbs) present or detected in both $B_{t_1}$ and $B_{t_2}$, denoted here as  $\mathbf{p}^k_1$ and $\mathbf{p}^k_2$, respectively. The joint $\mathbf{p}^k_i$ is obtained by finding the maximum value in the associated $k$-th heatmap of each bounding box $B_{t_i}$. For all parts detected both in $B_{t_1}$ and $B_{t_2}$, the color and texture content in a neighborhood around these two keypoints are compared. Let ${\cal{E}}$ be a squared neighborhood of 24$\times$24 pixels centered at $\mathbf{0}\in\mathbb{R}^2$.  Then,
    \begin{equation} 
        C_{c}(B_{t_1},B_{t_2})\!=\!\frac{1}{255 |S| \, |\cal{E}|}\!\!\sum_{k\in S}\!\sum_{\mathbf{y}\in{\cal{E}}}\!\|I_{t_1}(\mathbf{p}^k_1+\mathbf{y})-I_{t_2}(\mathbf{p}^k_2+\mathbf{y})\|
    \label{eq:ColEq}
   \end{equation}
where $S$ in Equation \eqref{eq:ColEq} denotes the set of mentioned pairs of corresponding  keypoints detected in both frames.
In the presented experiments, 
$\alpha=0.65$, $\beta=0.05$, $\gamma=0.3$.

%
\noindent To establish the matching assignments between bounding boxes along time, a memory criterion is used to set back on track those targets that have not been detected in particular single frames. For each frame $t$, the values of $C(B_{t},B_{t-1})$ and $C(B_{t},B_{t-2})$ are considered (for both frames' bounding boxes) and, by using a variant of the Hungarian algorithm, the bounding boxes of the current frame are assigned both to the ones detected in the $t-1$ and $t-2$. Those boxes that have only been detected in either $t-1$ or $t-2$ will be directly associated to the respective candidates, while the ones that have been properly found in both previous frames will be associated to the assignment that minimizes the overall matching cost. The output of this computation is an existing unique identifier for each matched bounding box, and a new identifier for boxes that could not be assigned.

\section{Results}
\label{sec:results}
This section focuses on the quantitative and qualitative evaluation of the proposed method. An ablation study assessing the contribution of each of the ingredients is included. \\ 
\noindent
\textbf{Dataset:} A dataset of 22 Basketball sequences has been built from several European basketball games corresponding to the \textit{Final Four} of the \textit{ANGT 2017}. The sequences have been directly extracted from the broadcasting video by manually setting the beginning and end times, hence avoiding having different camera shots; the original videos have full-HD resolution ($1920 \times 1080$) and 25 fps, but for the purposes of this experiment, only 4 frames are extracted per second, thus reducing computational expenses. The content of these sequences involve many different basketball offensive plays, such as isolation, \textit{pick and roll}, or even sideline strategies; besides, all 22 sequences include different jersey colors and different color-skinned players. The main limitation of this dataset is that it does not contain offensive transitions where players run from one side of the court to the other one, as camera stabilization is not able to handle this situation. The mean duration of these sequences is 11.07 seconds, resulting in a total of 1019 frames. A ground truth has been manually generated with the by dragging bounding boxes over each player and all 3 referees (taking the minimum visible X and Y coordinates of each individual) in every single frame (when visible), which results in a total of 11339 boxes. 

\noindent
\textbf{Quantitative results:} Quantitative assessment of both the proposed detection and tracking methods is provided in Tables \ref{dectTable} and \ref{trackTable}. 
Table \ref{dectTable} focuses on evaluating the proposed detection strategy; in this context, bounding boxes assignments are computed in each frame from scratch according to the maximum (and non-zero) IoU between ground truth boxes and current annotations, thus leading to TPs, FPs, FNs and TNs. Besides, a comparison with the state-of-the-art YOLO network \cite{redmon2016you} is given; for a fair comparison, only the \textit{person} detections within the court boundaries are kept. The tracking results are discussed in Table \ref{trackTable}, showing widely used error metrics:  MOTA, which takes into account false positives, misses and missmatches; and MOTP, which measures detection distances (details in \cite{bernardin2008evaluating}). To provide an ablation study of each of the contributions, results for the three stages described in Section \ref{subsec:playerdect} are included: the coarse-to-fine strategy of stages 1 and 1A, the sliding window method on the original (full-HD) video frames described in stage 2, and their combination. Moreover, tracking results have been compared with \cite{milan2015joint} as a State-of-the-Art method; in all tests, our detections have been used, thus starting off with the same conditions. It has to be mentioned that results obtained with the proposed tracking method were obtained only with CPU usage, whilst the \textit{Joint Tracking + Segmentation} method had to be run on a High Performance Cluster.
\begin{table}
\parbox{.45\linewidth}{
\centering
\begin{tabular}{|c|c|c|c|}
\hline
                     & \textbf{Precision} & \textbf{Recall} & \textbf{F1-Score} \\ \hline
Coarse-to-Fine       & \textbf{0.9959}    & 0.8095          & 0.8923            \\ \hline
High-Res. Scanning   & 0.9947             & 0.8109          & 0.8926            \\ \hline
YOLO                 & 0.8401             & \textbf{0.9426} & 0.8876            \\ \hline
Proposed Method & 0.9900             & 0.8563          & \textbf{0.9178}   \\ \hline
\end{tabular}
\caption{Detection performance: Average precision, recall and F1-Score of each strategy, over all sequences.}\label{dectTable}
}
\hfill
\parbox{.45\linewidth}{
\centering
\begin{tabular}{|c|c|c|c|}
\hline
\multicolumn{2}{|c|}{}             & \textbf{MOTA}   & \textbf{MOTP}   \\ \hline
\multirow{3}{*}{\STAB{\rotatebox[origin=c]{90}{no mem.}}} & Coarse-to-fine & 0.5837          & 0.6109          \\ \cline{2-4} 
                  & High-Res. Scanning   & 0.5711          & 0.5729          \\ \cline{2-4} 
                  & Proposed Combination    & 0.6237          & 0.6086          \\ \hline
\multirow{4}{*}{\STAB{\rotatebox[origin=c]{90}{mem.}}} & Coarse-to-fine & 0.6259          & 0.6110          \\ \cline{2-4} 
                  & High-Res. Scanning    & 0.6134          & 0.5871          \\ \cline{2-4} 
                  & \textit{Joint Track. + Segm.}    & \textbf{0.7142} & 0.3375          \\ \cline{2-4} 
                  & Proposed Combination    & 0.6704          & \textbf{0.6138} \\ \hline
\end{tabular}
\caption{Tracking and memory performance.} \label{trackTable}
}
\end{table}

\noindent On the one hand, Table \ref{dectTable} shows that the first two individual strategies have high precision values, but those suffer a drop in recall that can be compensated by merging detections. This improvement indicates that non-detected players (false negatives) are not the same ones when using the sliding window method than when implementing the coarse-to-fine approach; as mentioned,  the second method deals better with motion blur. Comparing with YOLO detections, it can be seen that lower recall is obtained, but it comes at a cost of precision; while our method does not detect some players, YOLO detects some non-existing players. This trade-off is compensated with the F1-score, which shows that the proposed method outperforms the SoA network. \\
On the other hand, tracking metric results in Table \ref{trackTable} lead to several conclusions. First, it is proved that introducing memory (with a tolerance of only 2 frames) in the matching procedure increases the MOTA metric by 5\%. Second, the combination of approaches results in less missed players/frame (less false negatives), thus providing better MOTA results. It can also be seen that the performance in terms of MOTP, which considers the IoU between matched instances, is better when using the full-HD scan technique. As seen in Figure \ref{fig:opBoxes}, the bounding boxes surrounding the human skeleton given by the pose model do not include the top-bottom parts of the human body, such as the forehead or feet; besides, the area of the bounding boxes depends on how many human parts have been found. As mentioned, the coarse-to-fine approach is better than the full-HD scanning at detecting partially occluded players, which are the most likely ones not to have all parts detected, thus resulting in notable area changes. Finally, better MOTA results are obtained with the \textit{Joint Tracking + Segmentation} method, but these come at a cost of MOTP for a main reason: their method performs segmentation inside every bounding box jointly with tracking, 
but our boxes contain really challenging situations, where players have random and stretched poses (plus occlusions), thus resulting in a poor box refinement and a MOTP drop. For this reason, the SoA method is an optimal tracker in constrained situations, where targets' bounding boxes contain almost no background and segmentation barely changes their size (\textit{i.e} scenes where the camera is further away than in basketball broadcasting sequences). 
Note that all performed tests have been done at 4 fps due to GT limitations; it is logic to hypothesize that with higher frame rates, MOTA results would improve due to closer proximity of targets, but it would come at a cost of computational expenses (and vice    versa with lower rates).

\section{Conclusions}\label{sec:conclusions}
In this article, a novel basketball player tracker based on multiple detections at different scales has been detailed. The proposed method is thought for single-camera systems and low-powered embedded devices, and uses -- once the court area is segmented -- a customized version of existing pose models, combining a coarse-to-fine approach and a sliding window technique at full resolution. Contextual and geometric features are then extracted for every single detection, and matched across consecutive frames. Having gathered a dataset from scratch, and having labelled thousands of ground truth bounding boxes, detection and tracking results can be independently obtained. Results are encouraging in both cases, proving that this technique could be used as the basis of a data analysis system for professional basketball teams. As future work, fast basketball transitions should be included in the dataset, and GPU-based pose models' performance should be tested with this multi-scale method together with other quantitative assessment tests; with more computational resources, deep learning features (\textit{i.e} output of a Convolutional Layer) could be used instead of contextual ones.

\section*{Acknowledgments}
The authors acknowledge partial support by MICINN/FEDER UE project, reference PGC2018-098625-B-I00, H2020-MSCA-RISE-2017 project, reference 777826 NoMADS and F.C. Barcelona's data support.

\end{document}